\documentclass{article} 

\usepackage{hyperref}
\usepackage{url}
\usepackage[utf8]{inputenc}
\usepackage{CJK}
\usepackage{amsmath}
\usepackage{amssymb}
\usepackage{amsthm}
\usepackage{tabularx}
\usepackage{graphicx}
\usepackage{float}
\usepackage{listings}
\usepackage{dsfont}
\usepackage{booktabs}
\usepackage{mathtools}
\usepackage{setspace}

\usepackage[accepted]{icml2019}

\begin{document}

\twocolumn[
\icmltitle{Combinational Q-Learning for Dou Di Zhu}

\icmlsetsymbol{equal}{*}

\begin{icmlauthorlist}
\icmlauthor{Yang You}{to}
\icmlauthor{Liangwei Li}{to}
\icmlauthor{Baisong Guo}{to}
\icmlauthor{Weiming Wang}{to}
\icmlauthor{Cewu Lu}{to}
\end{icmlauthorlist}

\icmlaffiliation{to}{Shanghai Jiao Tong University, Shanghai, China}

\icmlcorrespondingauthor{Cewu Lu}{lucewu@sjtu.edu.cn}


\vskip 0.3in
]



\printAffiliationsAndNotice{}  

\begin{abstract}
    Deep reinforcement learning (DRL) has gained a lot of attention in recent years, and has been proven to be able to play Atari games and Go at or above human levels. However, those games are assumed to have a small fixed number of actions and could be trained with a simple CNN network. In this paper, we study a special class of Asian popular card games called Dou Di Zhu, in which two adversarial groups of agents must consider numerous card combinations at each time step, leading to huge number of actions. We propose a novel method to handle combinatorial actions, which we call combinational Q-learning (CQL). We employ a two-stage network to reduce action space and also leverage order-invariant max-pooling operations to extract relationships between primitive actions. Results show that our method prevails over state-of-the art methods like naive Q-learning and A3C. We develop an easy-to-use card game environments and train all agents adversarially from sractch, with only knowledge of game rules and verify that our agents are comparative to humans. Our code to reproduce all reported results will be available online\footnote[2]{\url{https://github.com/qq456cvb/doudizhu-C}}.
\end{abstract}

\section{Introduction}
Recently, deep reinforcement learning has gained its advancement in games. AlphaGo \cite{silver2016mastering} first uses deep neural networks in board game Go to reduce the effective depth and breath of the search tree. AlphaGo efficiently combines the policy and value networks with Monte Carlo Tree Search (MCTS) and achieves superhuman performance in the game of Go. AlphaGo Zero \cite{silver2017mastering} is proposed and trained solely by self-play reinforcement learning, starting from random play, without any supervision or use of human data and it only uses only the black and white stones from the board as input features. In addition to board games, card games are a kind of games that also have an exponential number of states and are hard to solve. DeepStack \cite{moravvcik2017deepstack} is an algorithm that is able to solve Poker under imperfect information settings. It combines recursive reasoning to handle information asymmetry, decomposition to focus computation on the relevant decision, and a form of intuition that is automatically learned from self-play using deep learning. 

Though many games can be well solved by DRL, current DRL techniques, such as A3C \cite{mnih2016asynchronous} and double Q-learning \cite{van2016deep}, can not handle another card game called Dou Di Zhu. In this paper, we study Dou Di Zhu and explore a new solution to extent the ability of DRL. Dou Di Zhu is a popular game in China with a large number of players. In 2018, Tencent online game platform reported 30 million players attending annual Dou Di Zhu chaimpionship \cite{doudizhu-data3}.

There are three remarkable properties that make Dou Di Zhu totally different from previously mentioned board or card games. We list them as follows,
\begin{itemize}
    \item \textbf{Unconventional Representations.} The assumption of convolutional features in 2D board games and video games fails in Dou Di Zhu, since the knowledge lies in different combinations of cards at hand. 
    Therefore, we should introduce an unconventional representation for such kind of problem.  
    
    \item \textbf{Huge Action Space.} In Dou Di Zhu, the number of possible actions increases exponentially with the number of cards. At each round, a player needs to consider an action which is a subset of current handheld cards. Due to the complexity of Dou Di Zhu's game rule, there are a great variety of actions that one needs to consider and human players typically choose one valid action based on their rich experience and sometimes intuition.
    \item \textbf{Complicated Action Relationships.} The quality of each action depends largely on the conjunct influence of the cards to be handed out and those to be left.  One not only needs to consider the current action but also needs to consider what to give in the next several rounds. Thus relations between different cards needs to be taken in to consideration and this is what a human expert would do.   
\end{itemize}

To solve these challenges, we develop a two-stage hierarchical reinforcement learning  approach, which contains two parts: \textbf{Decomposition Proposal Network (DPN)} and \textbf{Move Proposal Network (MPN)}. During DPN, we choose the most promising decomposition based on its Q-value computed by order-invariant max-pooling operations; then during MPN, we pick up the final card group to be handed out. In addition, we random sample decompositions in DPN. Therefore, the dimension of action space at each level is considerably reduced and thus becomes computationally acceptable. Besides, we introduce special designed 1D-convolutional card representations which give enough information required by our networks.

In conclusion,  we propose a novel network architecture to handle combinational actions and show that it solves Dou Di Zhu by prevailing state-of-the-art methods like A3C and naive DQN \cite{mnih2013playing} and achieving human-level performance. We train three heterogeneous agents adversarially from scratch, without any domain knowledge except the game rules.

\section{Related Work}
Previous work on solving Dou Di Zhu \cite{whitehouse2011determinization} uses determinization to make decisions with stochasticity and imperfect information by sampling instances of the equivalent deterministic game of perfect information. They introduce a novel variant of MCTS that operates directly on trees of information sets. However, their performance is evaluated against relatively weak opponents and is not comparable to human players.

 Card games like Dou Di Zhu can be seen as a multi-step combinatorial bandits \cite{cesa2012combinatorial}, which is a combinatorial generalization of multi-step contextual bandits. Combinatorial generalizations of single-step contextual bandits \cite{cesa2012combinatorial, dani2008price} has been studied recently \cite{swaminathan2017off}. In their work, for each context (state), a policy selects a slate (action) consisting of component actions, after which a reward for the entire slate is observed. They also introduce a new practical estimator to evaluate a policy's performance. 
 
 Deep reinforcement learning with large discrete action spaces has also been studied \cite{dulac2015deep}. However, they strongly rely on prior information about the actions to embed them in a continuous space upon which their approach can generalize.

\section{Dou Di Zhu}
\label{sec:rule}
Dou Di Zhu \cite{wiki:Doudizhu} is a 3-player gambling card game, in the
class of climbing games but also with bidding elements similar to trick taking games. Dou Di Zhu originated in China, and has increased in popularity there in recent years, particularly with internet versions of the game. We adopt the rules from \cite{whitehouse2011determinization}.
\paragraph{Player Setting.} There are three players, \textit{Landlord}, \textit{Peasant Up}, \textit{Peasant Down}. During the game, players take their turns in a counterclockwise order; \textit{Peasant Up} denotes the player who plays right before \textit{Landlord} while \textit{Peasant Down} denotes the player who players right after \textit{Landlord}.

\paragraph{Card Deck.} A 3-player Dou Di Zhu uses a deck of 54 cards, which contains 15 different type of cards. These types are \{3, 4, 5, 6, 7, 8, 9, T, J, Q, K, A, 2, black joker, red joker\}, sorted by their ranks. There are four duplicate cards for each type, except for black joker and red joker.  At the beginning of the game, cards are randomly distributed to the three players and each player does not know others' cards.

\paragraph{Bidding Phase.}
Each player takes turns to bid on their hand with the possible bids being 1, 2 or 3 chips. Bids must be strictly higher than the current bid but each player has the option to pass. This continues until two of the players pass consecutively. If any player bids 3 chips then the bidding phase immediately ends. If all three players initially pass, the cards are shuffled and dealt again. The winner of the bidding phase is designated as the Landlord and this player adds the three extra cards on the table into their hand, and plays first. The winning bid determines the stake for the game.

\paragraph{Card Play Phase.}
The goal of the game is to be the first to get rid of all cards in hand. If the Landlord wins, the other two players must each pay the stake to the Landlord. However if either of the other two players wins, the Landlord pays the stake to both opponents. This means the two non- Landlord players must cooperate to beat the Landlord. The Landlord always plays first and then play moves around the table in a fixed direction. At the end of the game the stake is doubled if a player has failed to remove any cards from their hand.

The card play takes place in a number of rounds until a player
has no cards left. Whoever plays first can play any group of cards from their hand provided this group is a member of one of the legal move categories (see Table 1). The next player can play a group of cards from their hand provided this group is in the same category and has a higher rank than the group played by the previous player. If a player has no compatible group they must pass. This continues until two players pass, at which point the next player wins that round and may start a new round by playing a group of cards from any category.


Some categories allow extra kicker cards to be played with the
group which have no effect on the category or rank of the move being played. A kicker can be any card provided it is of different rank to all the cards in the main group. If the kicker cards are single cards they must be of different rank and if the kicker cards are pairs they must be differently ranked pairs. Also a Nuke cannot be used as a kicker. If a move with kickers is played, the next player must play a move in the same category with the same number of kickers, although the ranks of the kicker cards are ignored. 

\begin{tabularx}{0.48\textwidth}{|c|X|}
  \hline
  \textbf{Name} & \textbf{Description} \\
  \hline
  \textbf{Solo} & Any individual card, for example A or 2. It is also possible to play runs of sequential cards with length at least 5, for example 345678 or 89TJQKA. \\
  \hline
  \textbf{Pair} & Any pair of identically ranked cards for example 55 or 77. It is possible to play runs of sequential pairs with length at least 3, for example 334455 or TTJJQQKK. \\
  \hline
  \textbf{Trio} & Any three identically ranked cards for example AAA or 888. It is possible to play runs of sequential trios of any length, for example 444555 or TTTJJJQQQ. Each trio may also have a kicker attached, for example 444555TJ or 999QQ. \\
  \hline
  \textbf{Quadplex} & Any four identically ranked cards with two kickers attached, for examples 4444TJ or 999955KK. \\
  \hline
  \textbf{Bomb} & Any four identically ranked cards, for example 5555 or 2222. \\
  \hline
  \textbf{Nuke} & The red joker and the black joker together.\\
  \hline
\end{tabularx}

\section{Combinational Q-Learning in Dou Di Zhu}
\subsection{Stochastic Game with Imperfect Information}

Multi-Agent Reinforcement Learning can be defined under the framework of Stochastic Game \cite{van1981stochastic, yang2018mean}.
An N-agent stochastic game $G$ is expressed by a tuple $\langle \mathcal{S}, \mathcal{A}, p, r, \gamma\rangle$, where $\mathcal{S}$ denotes the state space and $\mathcal{A}$ is the joint action of all agents. Action space $\mathcal{A}$ can be factorized into each agent's action space $\mathcal{A}^j$, where $j = 1, \dots, N$ is the agent index. Likewise, $r$ is the reward function for all agents and can be factorized into $r^j:\mathcal{S} \times \mathcal{A} \to \mathbb{R}$. At each timestep, each agent takes an action $a^j\in \mathcal{A}^j$, forming a joint action $\mathbf{a}\in \mathcal{A} = \times_{\{j=1,\dots,N\}} \mathcal{A}^j$; then each agent receives a reward $r^j(s, \mathbf{a})$. State transition probabilities are defined by $p(s'|s,\mathbf{a}):\mathcal{S}\times \mathcal{A}\times \mathcal{S}\to [0, 1]$. $\gamma\in [0, 1]$ is the discount factor \cite{sutton2018reinforcement}.

The policy for each agent $j$ is $\pi^j:\mathcal{S}\to \Omega(\mathcal{A}^j)$ where $\Omega(\mathcal{A}^j)$ is the probability measure in space $\mathcal{A}^j$ and for finite dimension $dim(\mathcal{A}^j)$, $\Omega(\mathcal{A}^j)$ is just a simplex with dimension $dim(\mathcal{A}^j)-1$. $\pi = {\pi^1,\dots,\pi^N}$ denotes the joint policy of all $N$ agents. $\pi$ is often considered time-homogeneous, which means that it is independent of current timestep. Our aim is to maximize the value function for each agent:

\begin{align}
    V_\pi^j(s) = \sum_{t=0}^\infty\gamma^t\mathbb{E}_{\pi,p}[r_t^j|s_0=s;\pi].
\end{align}
Notice that it is a function of all agents' policy $\pi$ and state $s\in \mathcal{S}$.

We can then define the Q value as:
\begin{align}
    Q_\pi^j(s,\mathbf{a}) = r^j(s, \mathbf{a})+\gamma\mathbb{E}_{s'\sim p}[V_\pi^j(s')].
\end{align}
Note that this is a function of actions of all $N$ agents.

Notice that Dou Di Zhu is an imperfect information game where the full state $s$ (including all agents' handheld cards) is not observable to any individual so we resort to independent Q-learning \cite{tan1993multi}. In independent Q-learning, which is the simplest and most popular approach to multi-agent RL, each agent learns its own Q-function that conditions only on its observed state $s^j$ and its own action $a^j$. In deep RL, this is often done by having each agent perform deep Q-learning using the state and its own action. If we denote independent Q operator as $[\mathrm{Ind}Q^j]$ for each agent $j$,
\begin{align}
    [\mathrm{Ind}Q^j](s^j, a^j) \equiv r(s^j, a^j) + \mathbb{E}_{{s^j}'\sim p}[\gamma V^j({s^j}')].
\end{align}

Specifically, in Dou Di Zhu, each state $s^j$ corresponds to one's handheld cards and the cards handed out by the other two at this round. Besides, we augment the state with inferred handheld cards of the other two, which is done by calculating the distribution of remaining cards. Each action $a^j$ represents a legal move given the cards handed out by the other two. Positive one is given as rewards when the agent wins the game and negative one is given when it loses. For all other state-action pairs $(s^j,a^j)$, reward zero is given. For simplicity, we omit the bidding phase. In the following sections, we abuse the notation $s$ and $a$ for $s^j$ and $a^j$, ignoring their agent index.









\






\subsection{Combinational Q-Learning}
The original problem can be hard to solve and rarely converges due to our experiments in Section \ref{sec:soa}. The reason for this lies in the fact that there are over hundreds up to thousands of possible actions given one's handheld cards and standard Q-learning performs poorly on such a large combinational action space.

When considering a human playing Dou Di Zhu, it is common that human players tend to decompose their handheld cards according to the current situation. For example, when opponents hand ``A'' out as \textit{Solo}, a smart player would consider decomposing his cards if he holds two ``2'' in his hand. He would play ``2'' as \textit{Solo} instead of \textit{Pair} to take control.

This kind of decomposition also takes the relationship between card groups into consideration. Again consider some other player hands ``3'' out as \textit{Solo}, if a player holds three ``4'', it is not a good idea to split them to give ``4'' as \textit{Solo}. This is because, ``4'' is of rather small rank and leaving two ``4'' in hand is not a good choice; instead, making three ``4'' a \textit{Trio} with extra kicker cards is a more promising action. 

Inspired by how a human plays Dou Di Zhu, we employ a two-stage combinational Q-learning (CQL) algorithm that at each stage, only tens to a hundred actions need to be considered. For state $s$ in each agent's original Markov Decision Process (MDP), we replace it with two states, called $s_c$ and $s_f$ (``\textit{c}'' for ``\textit{combination}'' and ``\textit{f}'' for ``\textit{fine-grained action}''). When in $s_c$, agents choose the best decomposition and then in $s_f$, agents choose the best final move within previously selected decompisition. At each stage, a new set of actions need to be defined, $\mathcal{A}_c$ and $\mathcal{A}_f$ respectively. 

Denote current handheld cards as a set $\mathcal{H}$, all legal moves as $\mathcal{L}$, which is a set of card sets. $\mathcal{A}_c$ and $\mathcal{A}_f$ are defined as follows:
\begin{align}
    \mathcal{A}_c &\coloneqq \{ \mathcal{A}_f^{(1)}, \mathcal{A}_f^{(2)}, \dots, \mathcal{A}_f^{(D)} \} \\
    \mathcal{A}_f^{(i)} &\coloneqq   \{ C_{(i)}^1, C_{(i)}^2, \dots, C_{(i)}^K\}
\end{align}
where $D$ is the number of possible decompositions given current handheld cards and $K$ is the number of card groups within each decomposition. $C_{(i)}^j$ ($j=1,2,\dots,K$) is the card group to play at each round, described in section \ref{sec:rule}. To ensure that $\mathcal{A}_f^{(i)}$ is a valid decomposition, we need:
\begin{align}
    \cup_{j=1}^KC_{(i)}^j &= \mathcal{H},\\
    \cap_{j=1}^KC_{(i)}^j &= \O,\\
    C_{(i)}^j&\in \mathcal{L},\qquad \text{for all } j = 1,2,\dots,K
\end{align}

\begin{figure*}[ht]
\centering
\includegraphics[width=\textwidth]{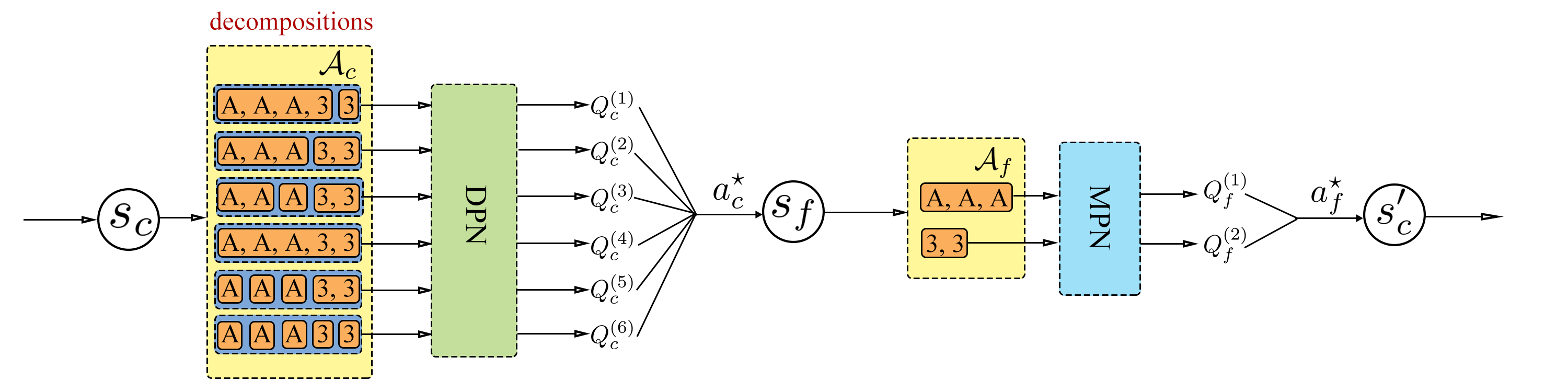}
\caption{\textbf{Augmented MDP in Dou Di Zhu.} When in $s_c$, agents choose the best decomposition among $\mathcal{A}_c$; then in $s_f$, agents choose the best final move among $\mathcal{A}_f$ within previously selected decompisition.}
\label{fig:mdp}
\end{figure*}

\begin{figure*}[ht]
\centering
\includegraphics[width=\textwidth]{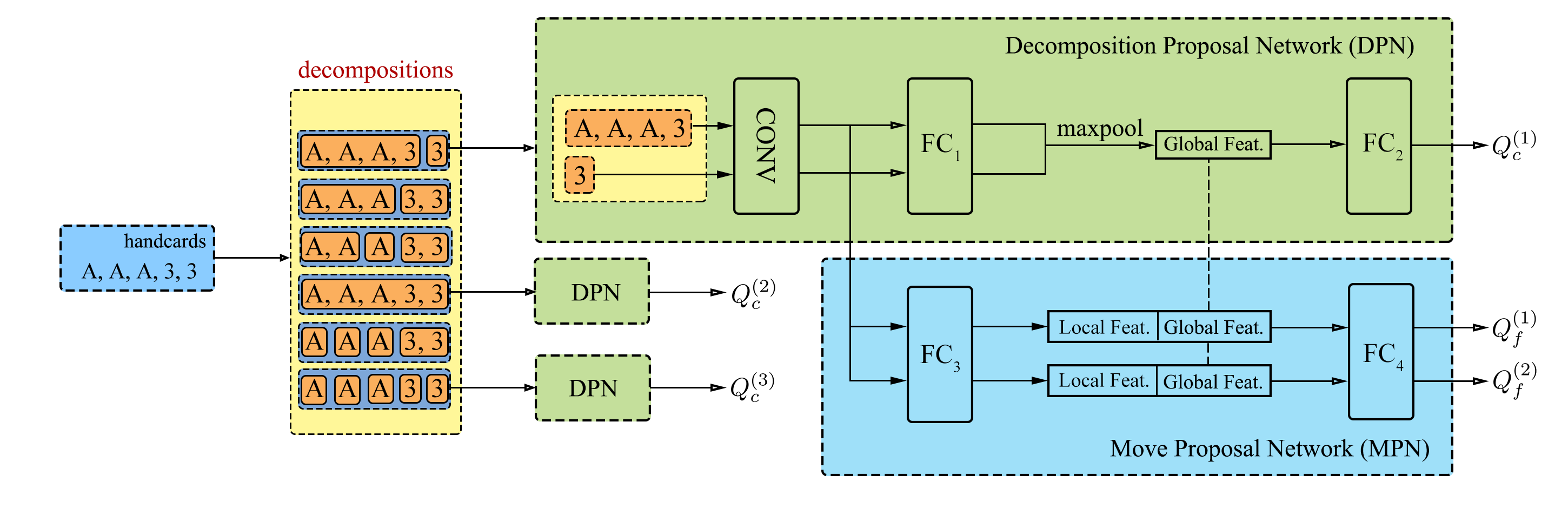}
\caption{\textbf{Our network structure of CQL on Dou Di Zhu.} We evaluate each decomposition with an order-invariant maxpooling operation in the end (DPN) and then each move's evaluation concatenates with this global feature (MPN). Networks in a single block share the same weights.}
\label{fig:network}
\end{figure*}

During online update of Q-learning, the original trajectory sample $(s, a, r)$ is replaced with two samples $(s_c, a_c, 0)$ and $(s_f, a_f, r)$, forming a two-stage hierarchical MDP, shown in Figure \ref{fig:mdp}. To this end, we design two novel networks: Decomposition Proposal Network (DPN) and Move Proposal Network (MPN), to evaluate corresponding Q values.
\paragraph{DPN.} In DPN, to get $Q_c^{(i)}\coloneqq Q(s_c,a_c^{(i)})$ where $a_c^{(i)} \coloneqq \mathcal{A}_f^{(i)}$, we adopt the idea of PointNet \cite{qi2017pointnet}. For each card group $C_{(i)}^j \subseteq \mathcal{H}$ represented as a 1D binary vector, we extract its 1D feature $f_{C_{(i)}^j}$ through 1D convolution layers (with average pooling in the end) followed by fully connected layers: $f_{C_{(i)}^j} = FC(CONV(C_{(i)}^j))$. Then we perform maxpooling on all card groups' features to get a global feature: $f_g^{(i)} = MAXPOOL(f_{C_{(i)}^1}, f_{C_{(i)}^2}, \dots,f_{C_{(i)}^K})$. Fully connected layers follow and $Q_c^{(i)} = FC(f_g^{(i)})$.
\paragraph{MPN.} After choosing the best decomposition $\mathcal{A}_f^{(i^\star)}$, in MPN, to get $Q_f^{(j)}\coloneqq Q(s_f,a_f^{(j)})$ where $a_f^{(j)} \coloneqq C_{(i^\star)}^j$, we concatenate each card group's local feature $f_{C_{(i^\star)}^j}$ with the global feature $f_g^{(i^\star)}$, passing it through fully connected layers: $Q_f^{(j)} = FC(CONCAT(f_{C_{(i^\star)}^j}, f_g^{(i^\star)}))$. The whole network architecture is shown in Figure \ref{fig:network}.

There are two advantages of employing this two-stage combinational Q-learning.

\begin{itemize}
    \item First, it greatly reduces the original action space into a hierarchical action space. In DPN, only the sampled decompositions need to be considered. Besides, after choosing the appropriate decomposition, MPN only considers each possible moves in this chosen decomposition.
    \item Secondly, in DPN, the relationship among all card groups (legal moves) in a single decomposition is also considered, analogy to human players' strategy on Dou Di Zhu. This relationship is extracted by a global order-invariant element-wise maxpooling, forming a global feature. This global feature is representative for the decomposition and can be used to get its Q value.
\end{itemize}

\section{Experiments}
In this section, we first describe the implementation details and then compare CQL with state-of-the-art reinforcement learning methods. Finally, we compare our self-play trained agents with other Dou Di Zhu baselines including human players. 

\subsection{Implementation Details}

\textbf{Fast Handheld Cards Decomposition.}
In our implementation, to find legal card groups that form a decomposition effectively, we leverage the method of dancing link \cite{knuth2000dancing}. 


Compared with naive depth-first-search algorithm, it dramatically reduces computational time of decomposing handheld cards (from tens of seconds to tens of milliseconds), especially when there are more than ten handheld cards. However, dancing link algorithm may miss some uncommon decompositions because of our ordered binary encoding scheme. Thus we employ dancing-link algorithm only when there are more than ten handheld cards and utilize naive depth-first-search otherwise. In practice, we find that this works well by keeping a balance between accuracy and efficiency.

\textbf{Card Group Auto-Encoder.}
For each card group $C_{(i)}^j \subseteq \mathcal{H}$, we extract its 1D feature through 1D convolution layers (\textit{CONV}), as shown in Figure \ref{fig:network}. Due to the special rule of Dou Di Zhu, we need to extract information for card group categories \textit{Solo}, \textit{Pair}, \textit{Trio}, \textit{Quadplex} respectively. Thus we build four convolutional layers that differ in kernel size and stride within \textit{CONV}.


We pre-train these convolutional layers as an auto-encoder. Variational Auto-Encoder \cite{doersch2016tutorial, kingma2013auto} is possible but considering there are only 13K+ fixed number of actions, a simple deterministic auto-encoder works well. 



\textbf{Networks.} In our experiments, we use double Q-learning \cite{van2016deep} and replay buffers \cite{schaul2015prioritized} to stabilize our learning progress. As shown in Figure \ref{fig:network}, for \textit{CONV}, we use eight ResNet residual blocks; for $FC_1$, we use three ResNet-like residual fully connected blocks with number of units 256, 512, 1024; for $FC_2$, $FC_3$, $FC_4$, we use three ResNet-like residual fully connected blocks with number of units 512, 256, 128. We refer readers to supplementary material for the full details of our network.
\par\textbf{Hyperparameters.} In our model, hyperparameters are chosen with Bayesian optimization together with memory and computational limits. Our chosen hyperparameters are shown in Table \ref{tab:hyper}.

\begin{table}[]
\centering
\begin{tabular}{|l|l|}
\hline
\textbf{Name}                                  & \textbf{Value} \\ \hline
Batch Size                            & 8     \\ \hline
Steps per Epoch                       & 2,500 \\ \hline
Update Frequency           & 4     \\ \hline
Memory Size                           & 3,000 \\ \hline
Random Sampling Size & 100   \\ \hline
\end{tabular}
\caption{\textbf{Hyperparameters in our experiments.} Steps per Epoch: number of parameter updates per epoch; Update Frequency: number of experience generation per network update; Memory Size: maximum size of replay buffer; Random Sampling Size: maximum number of samples of handheld card decompositions.}
\label{tab:hyper}
\end{table}

\subsection{Comparison to State-of-the-Art Methods}
\label{sec:soa}

In this section, we show that  how our proposed combinational Q-learning outperforms other baseline methods. 

\begin{figure}[ht]
\centering
\includegraphics[width=0.5\textwidth]{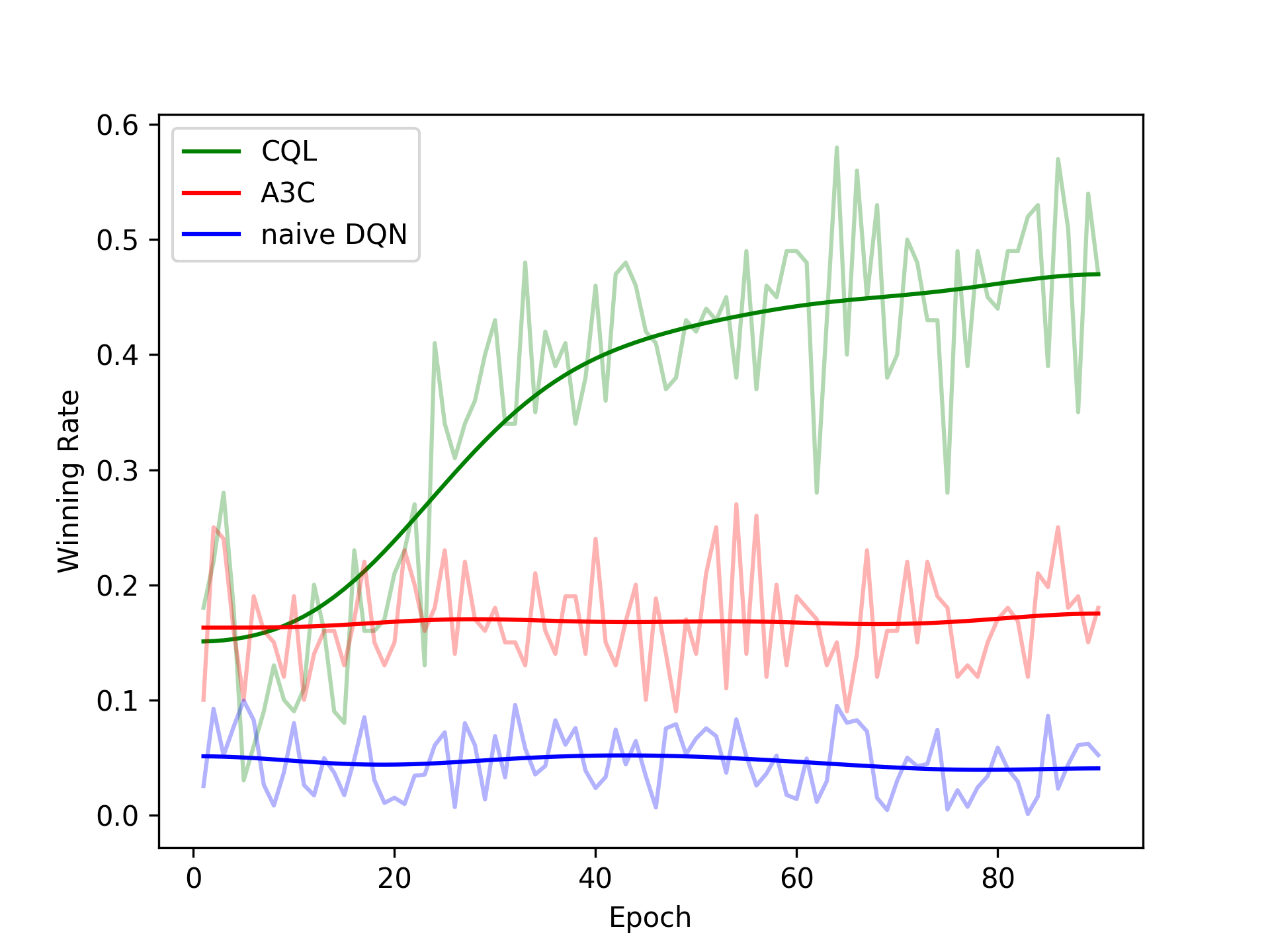}
\caption{\textbf{Winning rate of the agent \textit{Landlord} trained with combinational Q-learning, A3C and naive Q-learning methods.} All methods use RHCP as opponents (two \textit{Peasants}). \protect\footnotemark}
\label{fig:baseline}
\end{figure}

We train a single agent with a Recursive Handheld Cards Partitioning algorithm \cite{blog:DoudizhuAI} (RHCP, see details of this algorithm in supplementary material) as opponents. Here, we only train the agent \textit{Landlord}. We compare the winning rate of our algorithm with those of A3C \cite{mnih2016asynchronous} and naive DQN \cite{mnih2015human, van2016deep}. Combinational Q-learning shows a superior performance as shown in Figure \ref{fig:baseline}. The results are obtained by playing against the other two RHCP agents as environments for 50 episodes after each epoch.

\footnotetext{Naive Q-learning's gradient explodes up even with learning rate smaller than 1e-6 and we use the winning rate of the model with random assigned weights.}

From Figure \ref{fig:baseline}, we see that our proposed combinational Q-learning wins over all other baselines with a large gap (30\%). 

\textbf{Naive DQN} does not even converge since there are plenty of actions (more than 13K); the off-policy learning target with max-Q operation in Bellman equation becomes extremely unstable and overoptimistic given large discrete action spaces.


\textbf{A3C} works but only up to a limit. A3C introduces a value approximator and in some extent, it reduces variance introduced by large action spaces. However, it is still too hard for A3C to learn the special combinational structures of actions in card games like Dou Di Zhu.

\textbf{Combinational Q-learning} solves this problem with a huge improvement. It greatly raise the upper bound that could be obtained by deep learning approach. When utilizing combinational decompositions of handheld cards, training becomes much more stable.

\begin{figure*}[ht]
\centering
\includegraphics[width=1\textwidth]{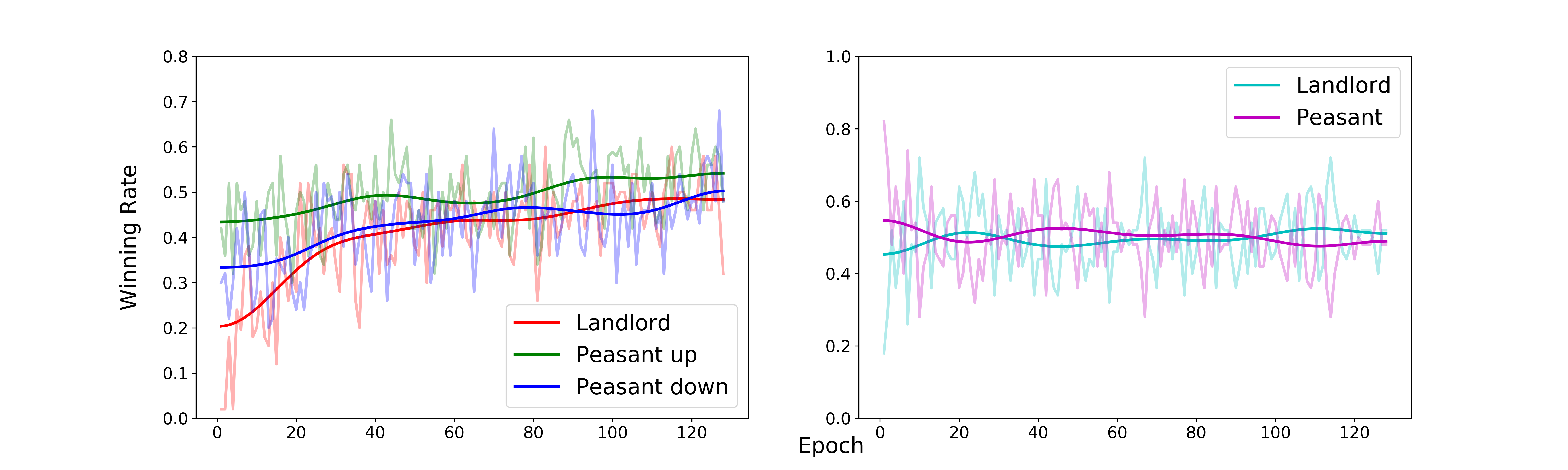}
\caption{\textbf{Learning curves for CQL.} \textit{Left:} Winning rates of different agents during training, evaluated using RHCP as opponents (environments) for 50 episodes after each epoch. \textit{Right:} Adversarial winning rates of \textit{Landlord} and \textit{Peasants} during training.}
\label{fig:magent}
\end{figure*}

\begin{figure*}[ht]
\centering
\includegraphics[width=0.8\textwidth]{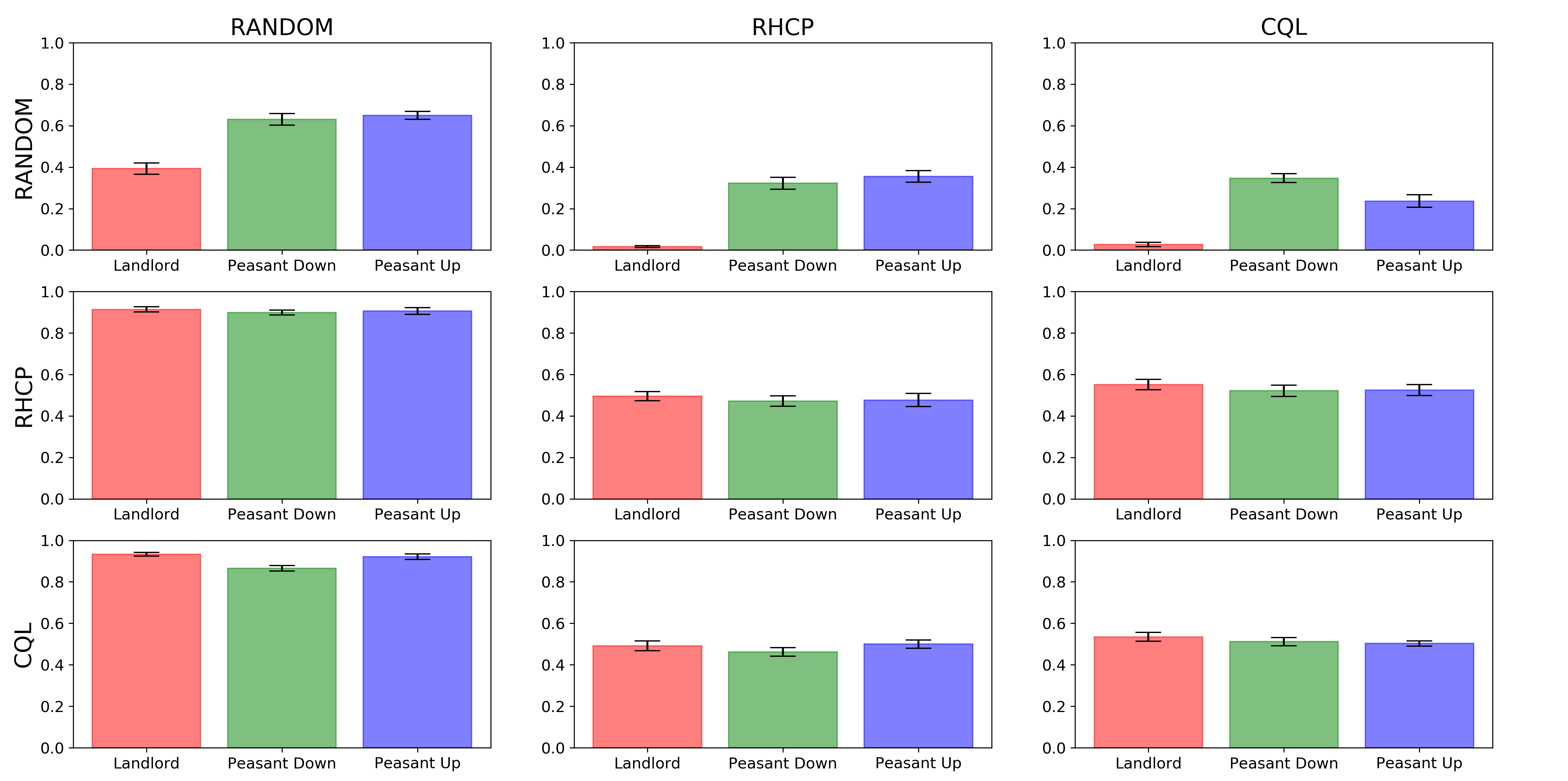}
\caption{\textbf{Winning rates of different models playing against each other.} \textit{Y-axis} denotes the type of model used as agent, which has three roles \textit{Landlord}, \textit{Peasant Down} and \textit{Peasant Up}. \textit{X-axis} denotes the type of model used as environments (the other two agents). For example, the left bottom figure shows the winning rate of our \{\textit{Landlord}, \textit{Peasant Down}, \textit{Peasant Up}\} player playing against the other two random players, respectively. Standard derivations are shown in black lines.}
\label{fig:compare33}
\end{figure*}

\subsection{Comparison to Dou Di Zhu Baselines}
\label{sec:a_mul}
There are three different roles in Dou Di Zhu namely \textit{Landlord} and two \textit{Peasants}. We utilize independent Q-learning in an asynchronous manner. To realize this, we train our agents simultaneously from scratch with only the information of game rules. During each training iteration, individuals behave in an environment with the other two agents as opponents. All parameters are updated online. We train the model on a server with one 32-core AMD Threadripper CPU and one 1080Ti GPU for 130 epochs in 20 days.

\paragraph{Learning Curve.}
Learning curve is shown in Figure \ref{fig:magent}.

From the bottom figure, we see that at first, \textit{Landlord} is rather weak and wins much less than \textit{Peasants}. However, through purely adversarial play, \textit{Landlord} becomes stronger quickly and can obtain a comparable winning rate with \textit{Peasants}. Throughout the whole training process, we see that \textit{Landlord} and \textit{Peasants} keep a balance in winning rates, meaning that our network is not likely to fall into local minima within which, one could easily defeat another.

From the top figure, we see that gradually, all three agents become stronger and stronger. \textit{Landlord} achieves the biggest improvement since \textit{Landlord} seldom wins under random actions. Besides \textit{Landlord}, the two \textit{Peasants} also learn from playing against \textit{Landlord} and obtain an improvement in learning rate of about five percent.

\paragraph{Performance Against RHCP and Random Agents.}
Since there is no ``oracle'' or public rankings for Dou Di Zhu, to evaluate our model against other baseline models (random, RHCP), we let them play against each other. For example, to test the performance of a single agent of our model, this agent would play against the other two random or RHCP based agents, which are considered as environments. All three agents' performance will be evaluated in this way and the results are shown in Figure \ref{fig:compare33}. The results are obtained by playing 100 episodes for 10 times with different random seeds.

We can see that CQL based agents achieve a comparable performance by playing against RHCP based agents. In contrast, we do not hard-code the conditions that agents may meet and our agents could potentially learn some patterns that beyond human's interpretation.
\begin{figure}[H]
\centering
\includegraphics[width=0.5\textwidth]{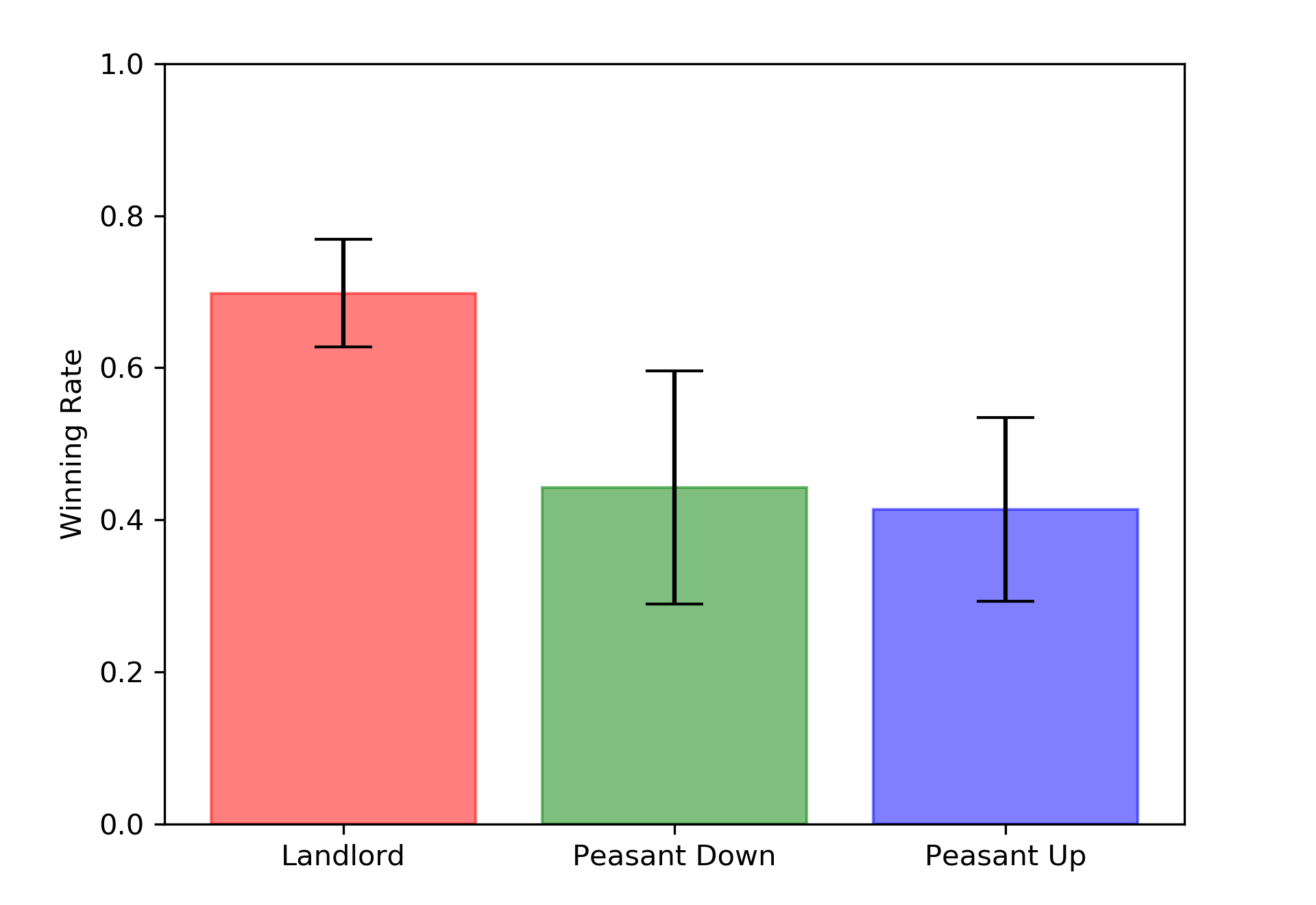}
\caption{\textbf{Winning rates of human players against our agents.} Different bars represent different roles of human players. Standard derivations are shown in black lines.}
\label{fig:human}
\end{figure}
\paragraph{Performance Against Humans}
We invite 20 people to play against our self-trained Dou Di Zhu agents. Each player plays about 35 games and the total number of games is 717. The winning rate of human players under different roles is shown in Figure \ref{fig:human}. 
We see that when human players being \textit{Landlord}, our agents have approximately 30\% winning rate; if human players are \textit{Peasants}, our agents could obtain 60\% winning rate. This indicates that our \textit{Landlord} agent is relatively stronger than two \textit{Peasant} agents and there exists a difficulty in cooperation between two \textit{Peasant} agents. In spite of the difficulty, we find that in some games, two \textit{Peasants} learn to cooperate without any supervision. For example, after \textit{Landlord} plays a card, \textit{Peasant Down} will hand out a big-rank card to take control, and then he hands out a small-rank card to let \textit{Peasant Up} hand out all his handheld cards in order to win. Table \ref{tab:humanaslord} visualizes one whole game process. More detailed game records can be found in supplemetary material.

\section{Conclusion}
In this paper, we introduce Dou Di Zhu as a challenging reinforcement learning environment. Then we propose a novel learning method combinational Q-learning (CQL) to handle combinational action space and compilcated action relationships in Dou Di Zhu. Our proposed DPN and MPN together not only outperforms other state-of-the-art reinforcement learning methods but also achieve a comparable performance against human players.

This work explores combinational action spaces in deep reinforcement learning and provides a new approach to solve it. While it solves the problem to a large extent, there are still potentials to further reduce action space by pruning actions based on some priors, given the current environment. It would also be of interest to explore the application of CQL in a more complex task or environment such as graph clustering or resource distribution.

Our proposed CQL also reveals some relations with hierarchical Q-learning. CQL can be seen as a special structured two-level hierarchical Q-learning with augmented MDPs. We leave investigation of this relationship as future work.

\begin{table*}[ht]
\centering
\begin{tabular}{@{}cccc@{}}
\toprule
No. Rounds & Role               & Current Hand Cards                       & Current Move   \\ \midrule
1          & Landlord: Human Player 1          & 3,4,4,4,4,5,6,6,7,8,9,T,J,Q,Q,A,A,A,2,2 & 5,6,7,8,9,T,J \\
1          & Peasant Down: CQL & 3,3,5,5,6,7,8,8,9,9,T,J,J,J,Q,K,K       & None           \\
1          & Peasant Up: CQL   & 3,5,6,7,7,8,9,T,T,Q,K,K,A,2,2,*,\$     & None           \\ \midrule
2          & Landlord: Human Player 1          & 3,4,4,4,4,6,Q,Q,A,A,A,2,2                & 3,4,4,4,4,6    \\
2          & Peasant Down: CQL & 3,3,5,5,6,7,8,8,9,9,T,J,J,J,Q,K,K       & None           \\
2          & Peasant Up: CQL   & 3,5,6,7,7,8,9,T,T,Q,K,K,A,2,2,*,\$     & None           \\ \midrule
3          & Landlord: Human Player 1          & Q,Q,A,A,A,2,2                            & Q,Q,A,A,A      \\
3          & Peasant Down: CQL & 3,3,5,5,6,7,8,8,9,9,T,J,J,J,Q,K,K       & None           \\
3          & Peasant Up: CQL   & 3,5,6,7,7,8,9,T,T,Q,K,K,A,2,2,*,\$     & *,\$           \\ \midrule
4          & Landlord: Human Player 1          & 2,2                                      & None           \\
4          & Peasant Down: CQL & 3,3,5,5,6,7,8,8,9,9,T,J,J,J,Q,K,K       & None           \\
4          & Peasant Up: CQL   & 3,5,6,7,7,8,9,T,T,Q,K,K,A,2,2          & 5,6,7,8,9      \\ \midrule
5          & Landlord: Human Player 1          & 2,2                                      & None           \\
5          & Peasant Down: CQL & 3,3,5,5,6,7,8,8,9,9,T,J,J,J,Q,K,K       & 6,7,8,9,T     \\
5          & Peasant Up: CQL   & 3,7,T,T,Q,K,K,A,2,2                    & None           \\ \midrule
6          & Landlord: Human Player 1          & 2,2                                      & None           \\
6          & Peasant Down: CQL & 3,3,5,5,8,9,J,J,J,Q,K,K                  & 5,5            \\
6          & Peasant Up: CQL   & 3,7,T,T,Q,K,K,A,2,2                    & 2,2            \\ \midrule
7          & Landlord: Human Player 1          & 2,2                                      & None           \\
7          & Peasant Down: CQL & 3,3,8,9,J,J,J,Q,K,K                      & None           \\
7          & Peasant Up: CQL   & 3,7,T,T,Q,K,K,A                        & T,T          \\ \midrule
8          & Landlord: Human Player 1          & 2,2                                      & 2,2           
\end{tabular}
\caption{\textbf{Game records played by \textit{Human Player 1} and CQL based agents.} \textit{Human Player 1} takes the role of \textit{Landlord} while the other two \textit{Peasants} are CQL based agents. \textit{Current Hand Cards} column denotes their handheld cards at the current round and \textit{Current Move} column denotes their handed out cards at this round. Note that "T" represents card "10", "*" represents "black joker" and "\$" represents "red joker". \textit{Human Player 1} wins.}

\label{tab:humanaslord}
\end{table*}







\clearpage
\clearpage

\bibliography{icml}
\bibliographystyle{icml2019}

\end{document}


\maketitle
\section{Recursive Hand Cards Partitioning (RHCP) Algorithm}

\begin{table}[]
    \label{tab:cat_score}
    \centering
        \begin{tabular}{|c|c|}
        \hline
        \textbf{Category}                         & \textbf{Weight}                \\ \hline
        None                             & 0                     \\ \hline
        Solo                             & MaxCard - 10          \\ \hline
        Pair                             & MaxCard - 10          \\ \hline
        Trio                             & MaxCard - 10          \\ \hline
        Sequential Solos                 & MaxCard - 10 + 1      \\ \hline
        Sequential Pairs                 & MaxCard - 10 + 1      \\ \hline
        Sequential Trios Take None       & MaxCard - 10 + 1      \\ \hline
        Sequential Trios Take One        & MaxCard - 10          \\ \hline
        Sequential Trios Take Two        & MaxCard - 10          \\ \hline
        Sequential Trios Series Take One & (MaxCard - 3 + 1) / 2 \\ \hline
        Sequential Trios Series Take Two & (MaxCard - 3 + 1) / 2 \\ \hline
        Bomb                             & MaxCard - 3 + 7       \\ \hline
        Four Take Two Solos              & (MaxCard - 3) / 2     \\ \hline
        Four Take Two Pairs              & (MaxCard - 3) / 2     \\ \hline
        Nuke                             & 20                    \\ \hline
        \end{tabular}
    \caption{\textbf{Definitions of the category scores of all the legal categories in RHCP Algorithm.} Here MaxCard represents different values in different categories. For a card group in the categories \textit{Solo}, \textit{Pair}, \textit{Trio}, \textit{Sequential Solos}, \textit{Sequential Pairs}, \textit{Sequential Trios Take None}, \textit{Bomb}, MaxCard denotes the max card value in the current card group (for example, the group $345678$ belongs to the category \textit{Sequential  Solos} and its MaxCard is $8$). For a card group in the categories \textit{Sequential Trios Take One}, \textit{Sequential Take Two}, \textit{Sequential Trios Series Take One}, \textit{Sequential Trios Series Take Two}, \textit{Four Take Two Solos}, \textit{Four Take Two Pairs}, MaxCard denotes the max card value of the principal cards(for example, the group QQQKKK89 belongs to the category \textit{Sequential Trios Series Take One}, and its principal cards are QQQKKK thus its MaxCard is that of QQQKKK which is $13$.}
\end{table}

\paragraph{General Idea.} The general idea of RHCP Algorithm is to take a best cards handing out strategy at each step. However, for a given set of hand cards, one is supposed to take all the possibilities into consideration and choose a best one among them, where RHCP takes effect. RHCP is inspired by the fact that any handing out strategy involves a certain way of partitioning current hand cards $\mathcal{H}$ into two card groups: $C$ and $\mathcal{H}\setminus C$, where $C$ denotes the card group that is to be handed out  and $\mathcal{H}\setminus C$ denotes the card group that is to be kept as remained hand cards. Thus what RHCP actually does is to pick a partitioning strategy with the highest Strategy Score which will be discussed next.

\paragraph{Strategy Score. }Strategy score $Q(C, \mathcal{H})$ is a function that measures the quality of handing out strategy $C$ given a set of hand cards $\mathcal{H}$. We formulate this strategy score function as:
\begin{equation}
    Q(C, \mathcal{H}) = 
    \begin{cases}
            r(C) + \max\limits_{C'\in \mathcal{H}}Q(C', \mathcal{H} \backslash C) & \mbox{if } \mathcal{H} \neq \emptyset \\
            0 & \mbox{if } \mathcal{H} = \emptyset
    \end{cases}
\end{equation}

where $r(C)$ is the category score for card group $C$ shown in Table \ref{tab:cat_score}. 
We select the best card group $C^\star$ given current hand cards $\mathcal{H}$ by,
\begin{align}
     C^\star = \argmax_C Q(C, \mathcal{H})
\end{align}

\section{A3C and naive DQN Network Details}

\begin{table}[]
\label{tab:hyper}
\centering
\begin{tabular}{|l|l|}
\hline
\textbf{Name}                                  & \textbf{Value} \\ \hline
Batch Size                            & 8     \\ \hline
Steps per Epoch                       & 2,500 \\ \hline
Update Frequency           & 4     \\ \hline
Memory Size                           & 3,000 \\ \hline
\end{tabular}
\caption{\textbf{Hyperparameters of A3C and naive DQN.}}
\end{table}
To verify the advantages of our model, we compare three different reinforcement learning architectures, which include naive Q-learning, Asynchronous Advantage Actor-Critic and our combinational Q-learning. This section describes the experiment details in this comparative experiment. What should be noticed is that here we all use hand-card-weight-based algorithm to be the opponent in both training or test.

\subsection{Hyperparameters}
All the models share the same hyperparameters as Combinational Q-learning to show a fair comparison. The parameters are shown in the Table \ref{tab:hyper}.

\subsection{Architecture Details}
\begin{figure}[ht]
\centering
\includegraphics[width=0.5\textwidth]{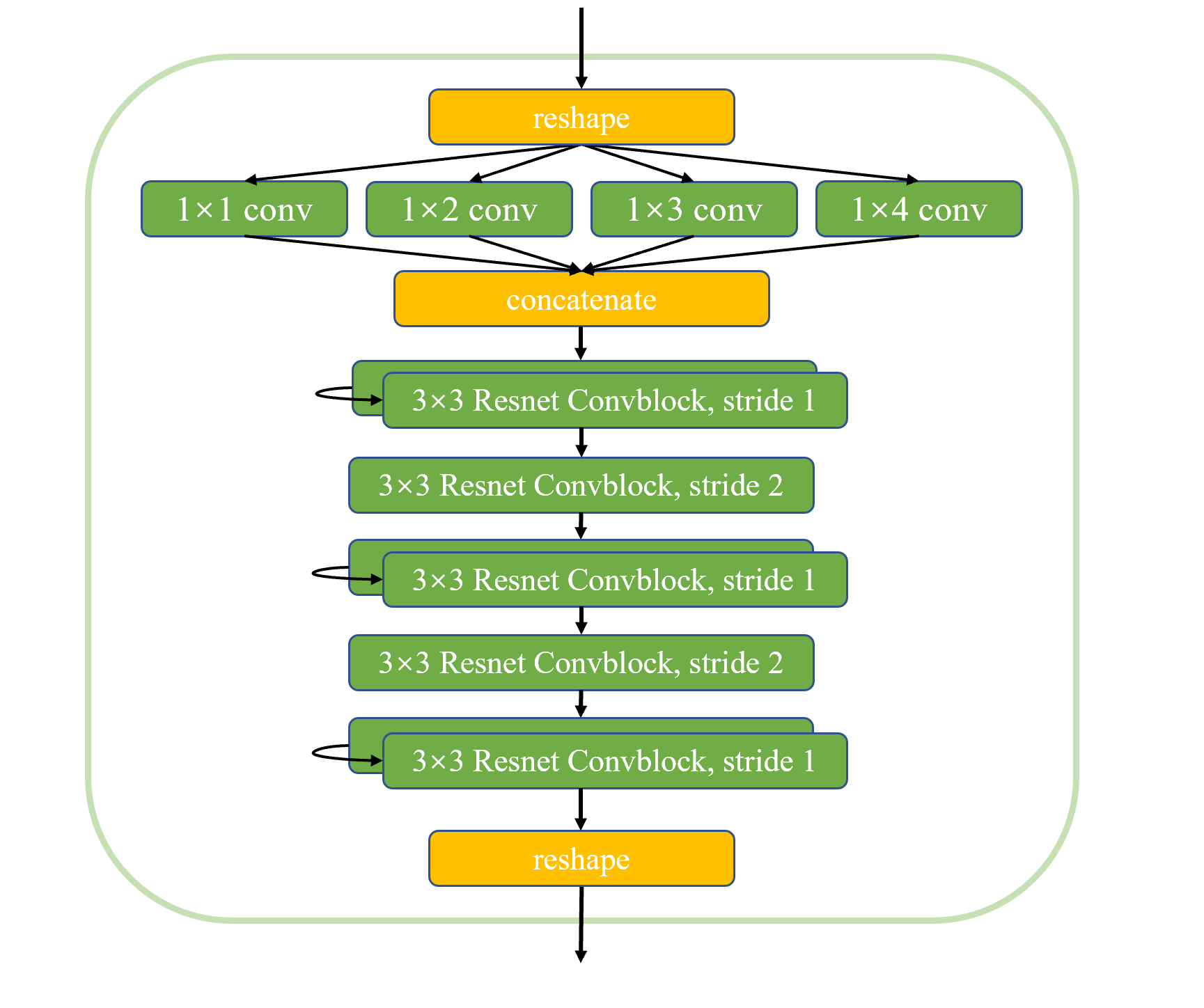}
\caption{\textbf{\textit{CONV} structures.}}
\label{fig:1dfeature}
\end{figure}

\paragraph{Naive Q-learning.} The input consists of two parts. The first part is \textit{all agents' cards}, which is a 180-dimensional tensor. Its first 60 dimension is the one-hot representation of the training agent's current hand cards while the middle 60 dimension represents the probability of possible cards in the previous player's hand and the last 60 is the probability of possible cards in the next player's hand. These probabilities can be inferred given histories of all agents. The second part is \textit{cards in the last round}, which is a 256 * 2 = 512 dimenstional tensor. It consists of card group encoding of the cards played by the other two players in the last round.

\begin{figure}[ht]
\centering
\includegraphics[width=0.25\textwidth]{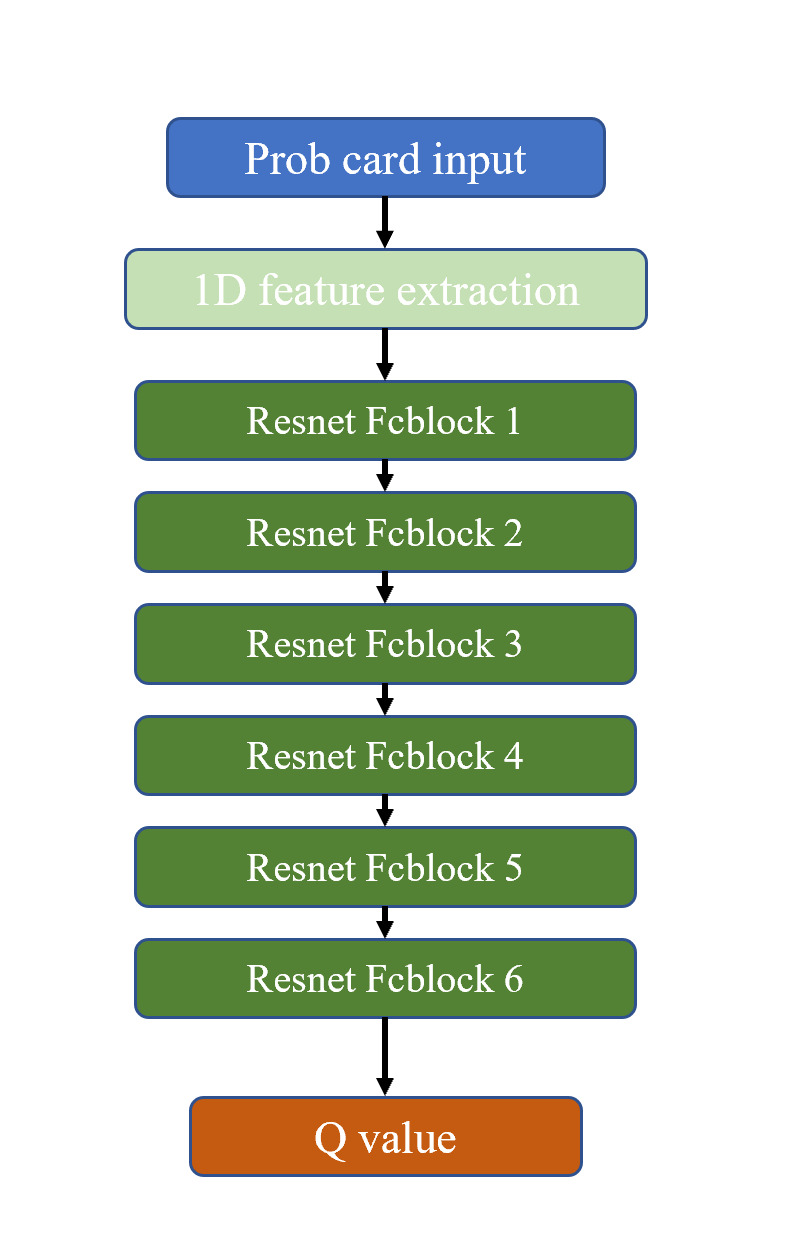}
\caption{\textbf{Naive Q-learning network structures.}}
\label{fig:naive}
\end{figure}

Then \textit{all agents' cards} passes through a special designed convolutional layers \textit{CONV} shown in Figure \ref{fig:1dfeature}, also discussed in Section 5.1. Then it is concatenated with \textit{cards in the last round}. A few fully connected layers follow and we get the output Q values for each action. The output is a 13527-dimensional tensor giving the values of all the possible action including the illegal operation which will be masked during inference. The network architecture is shown in Figure \ref{fig:naive}.

\paragraph{A3C.} We describe A3C network in two parts: the policy network and the value network. 

The input of policy network is nearly identical to that of naive Q-learning except for that instead of using a fixed pretrained auto encoder, we allow its paramters to update during training. The input will then pass through a LSTM with 1024 units. A few fully connected layers follow and we output action scores for both \textit{passive} and \textit{active} modes. In \textit{passive} mode, a player is required to respond to other players cards while in \textit{active} mode, a player takes control and decides what to give next. The action scores are 13527-dimensional probability distribution.  

\begin{figure}[h]
\centering
\includegraphics[width=0.5\textwidth]{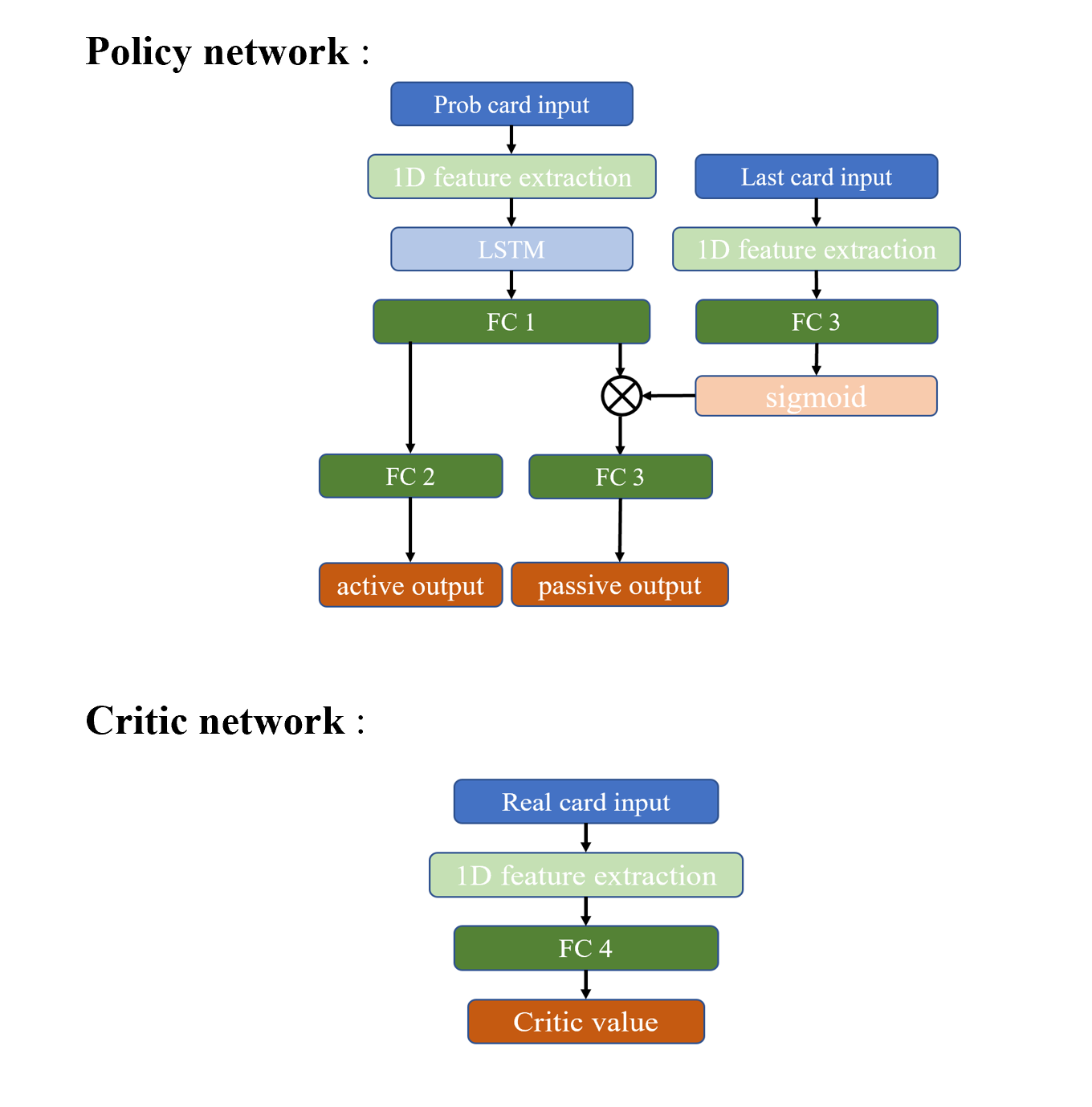}
\caption{\textbf{A3C network structures.}}
\label{fig:A3C}
\end{figure}

In the value network, since it's not used during inference, we feed all agents' cards instead inferring the other agents' cards to the network. Then similar special-designed convolutional layers as naive Q-learning apply and the output is passed through several fully-connected layers to get a final value. The network architecture is shown in Figure \ref{fig:A3C}.

\begin{table*}[h!]
\centering
\begin{tabular}{@{}cccc@{}}
\toprule
No. Rounds & Role                   & Current Hand Cards                         & Current Move \\ \midrule
1          & Landlord: CQL          & 3,4,5,5,6,6,7,8,T,T,T,J,Q,Q,K,A,A,2,2,* & 3,4,5,6,7,8  \\
1          & Peasant Down: CQL      & 3,4,4,4,5,7,7,8,8,T,J,J,K,A,A,2,\$        & None         \\
1          & Peasant Up:Human Player 2 & 3,3,5,6,6,7,8,9,9,9,9,J,Q,Q,K,K,2          & None         \\ \midrule
2          & Landlord: CQL          & 5,6,T,T,T,J,Q,Q,K,A,A,2,2,*             & 5,T,T,T   \\
2          & Peasant Down: CQL      & 3,4,4,4,5,7,7,8,8,T,J,J,K,A,A,2,\$        & None         \\
2          & Peasant Up:Human Player 2 & 3,5,6,6,7,8,9,9,9,9,J,Q,Q,K,K,2            & None         \\ \midrule
3          & Landlord: CQL          & 6,J,Q,Q,K,A,A,2,2,*                        & 6            \\
3          & Peasant Down: CQL      & 3,4,4,4,5,7,7,8,8,T,J,J,K,A,A,2,\$        & 7            \\
3          & Peasant Up:Human Player 2 & 3,3,5,6,6,7,8,9,9,9,9,J,Q,Q,K,K,2          & 8            \\ \midrule
4          & Landlord: CQL          & J,Q,Q,K,A,A,2,2,*                          & J            \\
4          & Peasant Down: CQL      & 3,4,4,4,5,7,8,8,T,J,J,K,A,A,2,\$          & None         \\
4          & Peasant Up:Human Player 2 & 3,3,5,6,6,7,9,9,9,9,J,Q,Q,K,K,2            & None         \\ \midrule
5          & Landlord: CQL          & Q,Q,K,A,A,2,2,*                            & Q,Q          \\
5          & Peasant Down: CQL      & 3,4,4,4,5,7,8,8,T,J,J,K,A,A,2,\$          & A,A          \\
5          & Peasant Up:Human Player 2 & 3,3,5,6,6,7,9,9,9,9,J,Q,Q,K,K,2            & None         \\ \midrule
6          & Landlord: CQL          & K,A,A,2,2,*                                & 2,2          \\
6          & Peasant Down: CQL      & 3,4,4,4,5,7,8,8,T,J,J,K,2,\$              & None         \\
6          & Peasant Up: Human Player 2        & 3,3,5,6,6,7,9,9,9,9,J,Q,Q,K,K,2            & None         \\ \midrule
7          & Landlord: CQL          & K,A,A,*                                    & K            \\
7          & Peasant Down: CQL      & 3,4,4,4,5,7,8,8,T,J,J,K,2,\$              & 2            \\
7          & Peasant Up:Human Player 2 & 3,3,5,6,6,7,9,9,9,9,J,Q,Q,K,K,2            & None         \\ \midrule
8          & Landlord: CQL          & A,A,*                                      & *            \\
8          & Peasant Down: CQL      & 3,4,4,4,5,7,8,8,T,J,J,K,\$                & \$           \\
8          & Peasant Up:Human Player 2 & 3,3,5,6,6,7,9,9,9,9,J,Q,Q,K,K,2            & None         \\ \midrule
9          & Landlord: CQL          & A,A                                        & None         \\
9          & Peasant Down: CQL      & 3,4,4,4,5,7,8,8,T,J,J,K                   & 4,4,4,8,8    \\
9          & Peasant Up:Human Player 2 & 3,3,5,6,6,7,9,9,9,9,J,Q,Q,K,K,2            & None         \\ \midrule
10         & Landlord: CQL          & A,A                                        & None         \\
10         & Peasant Down: CQL      & 3,5,7,T,J,J,K                             & J,J          \\
10         & Peasant Up:Human Player 2 & 3,3,5,6,6,7,9,9,9,9,J,Q,Q,K,K,2            & Q,Q          \\ \midrule
11         & Landlord: CQL          & A,A                                        & A,A         
\end{tabular}
\caption{\textbf{Game records played by \textit{Human Player 2} and CQL based agents.} \textit{Human Player 2} takes the role of \textit{Peasant Up} while \textit{Peasant Down} and \textit{Landlord} are CQL based agents. \textit{Current Hand Cards} column denotes their handheld cards at the current round and \textit{Current Move} column denotes their handed out cards at this round. Note that "T" represents card "10", "*" represents "black joker" and "\$" represents "red joker". CQL based \textit{Landlord} wins.}
\label{tab:humanasup}
\end{table*}

\begin{table*}[ht]
\centering
\begin{tabular}{@{}cccc@{}}
\toprule
No. Rounds & Role                    & Current Hand Cards                      & Current Move   \\ \midrule
1          & Landlord: CQL           & 3,3,4,4,5,6,7,7,8,9,J,J,Q,K,K,K,A,A,2,* & 3,4,5,6,7,8,9  \\
1          & Peasant Down: Human Player 3 & 5,5,6,6,7,8,9,9,9,T,T,T,J,K,2,2,\$   & 5,6,7,8,9,T,J \\
1          & Peasant Up: CQL         & 3,3,4,4,5,6,7,8,8,T,J,Q,Q,Q,A,A,2      & None           \\ \midrule
2          & Landlord: CQL           & 3,4,7,J,J,Q,K,K,K,A,A,2,*               & None           \\
2          & Peasant Down: Human Player 3 & 5,6,9,9,T,T,K,2,2,\$                  & 5              \\
2          & Peasant Up: CQL         & 3,3,4,4,5,6,7,8,8,T,J,Q,Q,Q,A,A,2      & None           \\ \midrule
3          & Landlord: CQL           & 3,4,7,J,J,Q,K,K,K,A,A,2,*               & Q              \\
3          & Peasant Down: Human Player 3 & 6,9,9,T,T,K,2,2,\$                    & K              \\
3          & Peasant Up: CQL         & 3,3,4,4,5,6,7,8,8,T,J,Q,Q,Q,A,A,2      & None           \\ \midrule
4          & Landlord: CQL           & J,Q,Q,K,A,A,2,2,*                       & J              \\
4          & Peasant Down: Human Player 3 & 3,4,4,4,5,7,8,8,T,J,J,K,A,A,2,\$       & None           \\
4          & Peasant Up: CQL         & 3,3,5,6,6,7,9,9,9,9,J,Q,Q,K,K,2         & None           \\ \midrule
5          & Landlord: CQL           & 3,4,7,J,J,K,K,K,A,A,2,*                 & *              \\
5          & Peasant Down: Human Player 3 & 6,9,9,T,T,2,2,\$                      & \$             \\
5          & Peasant Up: CQL         & 3,3,4,4,5,6,7,8,8,T,J,Q,Q,Q,A,A,2      & None           \\ \midrule
6          & Landlord: CQL           & 3,4,7,J,J,K,K,K,A,A,2                   & None           \\
6          & Peasant Down: Human Player 3 & 3,4,4,4,5,7,8,8,T,J,J,K,2,\$           & None           \\
6          & Peasant Up: CQL         & 3,3,5,6,6,7,9,9,9,9,J,Q,Q,K,K,2         & None           \\ \midrule
7          & Landlord: CQL           & K,A,A,*                                 & K              \\
7          & Peasant Down: Human Player 3 & 6,9,9,T,T,2,2                         & 9,9            \\
7          & Peasant Up: CQL         & 3,3,4,4,5,6,7,8,8,T,J,Q,Q,Q,A,A,2      & None           \\ \midrule
8          & Landlord: CQL           & 3,4,7,J,J,K,K,K,A,A,2                   & J,J            \\
8          & Peasant Down: Human Player 3 & 6,T,T,2,2                             & 2,2            \\
8          & Peasant Up: CQL         & 3,3,4,4,5,6,7,8,8,T,J,Q,Q,Q,A,A,2      & None           \\ \midrule
9          & Landlord: CQL           & 3,4,7,K,K,K,A,A,2                       & None           \\
9          & Peasant Down: Human Player 3 & 6,T,T                                 & T,T          \\
9          & Peasant Up: CQL         & 3,3,4,4,5,6,7,8,8,T,J,Q,Q,Q,A,A,2      & None           \\ \midrule
10         & Landlord: CQL           & 3,4,7,K,K,K,A,A,2                       & A,A            \\
10         & Peasant Down: Human Player 3 & 6                                       & None           \\
10         & Peasant Up: CQL         & 3,3,4,4,5,6,7,8,8,T,J,Q,Q,Q,A,A,2      & None           \\ \midrule
11         & Landlord: CQL           & 3,4,7,K,K,K,2                           & 3,K,K,K        \\
11         & Peasant Down: Human Player 3 & 6                                       & None           \\
11         & Peasant Up: CQL         & 3,3,4,4,5,6,7,8,8,T,J,Q,Q,Q,A,A,2      & None           \\ \midrule
 \end{tabular}
\caption{\textbf{Game records played by \textit{Human Player 3} and CQL based agents (first part).} \textit{Human Player 3} takes the role of \textit{Peasant Down} while \textit{Peasant up} and \textit{Landlord} are CQL based agents. \textit{Current Hand Cards} column denotes their handheld cards at the current round and \textit{Current Move} column denotes their handed out cards at this round. Note that "T" represents card "10", "*" represents "black joker" and "\$" represents "red joker". CQL based \textit{Landlord} wins.}
\label{tab:humanasdown1}
\end{table*}

\begin{table*}[t]
\centering
\begin{tabular}{@{}cccc@{}}
\toprule
No. Rounds & Role                    & Current Hand Cards                      & Current Move   \\ \midrule
12         & Landlord: CQL           & 4,7,2                                   & 2              \\
12         & Peasant Down: Human Player 3 & 6                                       & None           \\
12         & Peasant Up: CQL         & 3,3,4,4,5,6,7,8,8,T,J,Q,Q,Q,A,A,2      & None           \\ \midrule
13         & Landlord: CQL           & 4,7                                     & 7              \\
13         & Peasant Down: Human Player 3 & 6                                       & None           \\
13         & Peasant Up: CQL         & 3,3,4,4,5,6,7,8,8,T,J,Q,Q,Q,A,A,2      & 2              \\ \midrule
14         & Landlord: CQL           & 4                                       & None           \\
14         & Peasant Down: Human Player 3 & 6                                       & None           \\
14         & Peasant Up: CQL         & 3,3,4,4,5,6,7,8,8,T,J,Q,Q,Q,A,A        & 3,4,5,6,7      \\ \midrule
15         & Landlord: CQL           & 4                                       & None           \\
15         & Peasant Down: Human Player 3 & 6                                       & None           \\
15         & Peasant Up: CQL         & 3,4,8,8,T,J,Q,Q,Q,A,A                  & T,Q,Q,Q       \\ \midrule
16         & Landlord: CQL           & 4                                       & None           \\
16         & Peasant Down: Human Player 3 & 6                                       & None           \\
16         & Peasant Up: CQL         & 3,4,8,8,J,A,A                           & A,A            \\ \midrule
17         & Landlord: CQL           & 4                                       & None           \\
17         & Peasant Down: Human Player 3 & 6                                       & None           \\
17         & Peasant Up: CQL         & 3,4,8,8,J                               & 8,8            \\ \midrule
18         & Landlord: CQL           & 4                                       & None           \\
18         & Peasant Down: Human Player 3 & 6                                       & None           \\
18         & Peasant Up: CQL         & 3,4,J                                   & J              \\ \midrule
19         & Landlord: CQL           & 4                                       & None           \\
19         & Peasant Down: Human Player 3 & 6                                       & None           \\
19         & Peasant Up: CQL         & 3,4                                     & 3              \\ \midrule
20         & Landlord: CQL           & 4                                       & 4             
\end{tabular}
\caption{\textbf{Game records played by \textit{Human Player 3} and CQL based agents (second part).} \textit{Human Player 3} takes the role of \textit{Peasant Down} while \textit{Peasant up} and \textit{Landlord} are CQL based agents. \textit{Current Hand Cards} column denotes their handheld cards at the current round and \textit{Current Move} column denotes their handed out cards at this round. Note that "T" represents card "10", "*" represents "black joker" and "\$" represents "red joker". CQL based \textit{Landlord} wins.}
\label{tab:humanasdown2}
\end{table*}

\section{Detailed Game Records Against Human Players}
In this section, we present two other detailed game records as shown in Table \ref{tab:humanasup}, Table \ref{tab:humanasdown1} and Table \ref{tab:humanasdown2}.